\def\year{2019}\relax
\documentclass[letterpaper]{article} 

\usepackage{aaai19}  
\usepackage{times}  
\usepackage{helvet}  
\usepackage{courier}  
\usepackage{url}  
\usepackage{graphicx}  
\usepackage[nolist]{acronym}
\usepackage{amsmath}
\usepackage{amsfonts}
\usepackage{multirow}
\usepackage{subfig}

\title{Training Temporal Word Embeddings with a Compass}
\author{
Valerio Di Carlo,\textsuperscript{\rm 1}
Federico Bianchi,\textsuperscript{\rm 2}
Matteo Palmonari\textsuperscript{\rm 2}\\
\textsuperscript{\rm 1}BUP Solutions, Rome, Italy,
\textsuperscript{\rm 2}University of Milan-Bicocca, Milan, Italy\\
valerio.dicarlo@bupsolutions.com, 
federico.bianchi@disco.unimib.it,
palmonari@disco.unimib.it}

\begin{document}
%

%
\maketitle
\begin{abstract}
Temporal word embeddings have been proposed to support the analysis of word meaning shifts during time and to study the evolution of languages. Different approaches have been proposed to generate vector representations of words that embed their meaning during a specific time interval. 
However, the training process used in these approaches is complex, may be inefficient or it may require large text corpora. As a consequence, these approaches may be difficult to apply in resource-scarce domains or by scientists with limited in-depth knowledge of embedding models.  In this paper, we propose a new heuristic to train temporal word embeddings based on the Word2vec model. The heuristic consists in using atemporal vectors as a reference, i.e., as a \textit{compass}, when training the representations specific to a given time interval. The use of the \textit{compass} simplifies the training process and makes it more efficient. Experiments conducted using state-of-the-art datasets and methodologies suggest that our approach outperforms or equals comparable approaches while being more robust in terms of the required corpus size.
\end{abstract}

\begin{acronym}
\acro{atmodel}[TWEC]{Temporal Word Embeddings with a Compass}
\acro{twem}[TWEM]{Temporal Word Embedding Model}
\acro{twe}[TWE]{Temporal Word Embedding}
\acro{datas}[NAC-S]{News Article Corpus Small}
\acro{datab}[NAC-L]{News Article Corpus Large}
\acro{dataml}[MLPC]{Machine Learning Papers Corpus}
\acro{tests}[T1]{Testset1}
\acro{testb}[T2]{Testset2}
\acro{wem}[WEM]{Word Embedding Model}
\acro{twa}[TWA]{Temporal Word Analogy}
\end{acronym}

\section{Introduction} \label{introduction}
Language is constantly evolving, reflecting the continuous changes in the world and the needs of the speakers. While new words are introduced to refer to new concepts and experiences (e.g., \textit{Internet}, \textit{hashtag}, \textit{microaggression}), some words are subject to semantic shifts, i.e., their meanings change over time \cite{Basile2016DiachronicNgram}. For example, in the English language, the word \textit{gay} originally meant \textit{joyful}, \textit{happy}; only during the $20$th century the word began to be used in association with sexual orientation \cite{Kulkarni2015StatisticallyChange}. 

Finding methods to represent and analyze word evolution over time is a key task to understand the dynamics of human language, revealing statistical laws of semantic evolution \cite{Hamilton2016DiachronicChange}. In addition, time-dependent word representations may be useful when natural language processing algorithms that use these representations, e.g., an entity linking algorithm \cite{YamadaTACL1065}, are applied to texts written in a specific time period. 

\textit{Distributional semantics}  advocates a ``usage-based'' perspective on word meaning representation: this approach is based on the \textit{distributional hypothesis} \cite{Firth1957A1930-1955}, which states that the meaning of a word can be defined by the word's \textit{context}.

Word embedding models based on this hypothesis have received great attention over the last few years, driven by the success of the neural network-based model \textit{Word2vec} \cite{Mikolov2013EfficientSpace}. These models represent word meanings as vectors, i.e., \textit{word embeddings}.
Most state-of-the-art approaches, including Word2vec, are formulated as static models. Since they assume that the meaning of each word is fixed in time, they do not account for the semantic shifts of words. Thus, recent  approaches have tried to capture the dynamics of language 
\cite{Hamilton2016DiachronicChange,Bamler2017DynamicEmbeddings,Szymanski2017TemporalEmbeddings,Yao2017DiscoveryLearning,Rudolph2017DynamicEvolution}.

A \ac{twem} is a model that learns \textit{temporal word embeddings}, i.e., vectors that represent the meaning of words during a specific temporal interval. For example, a \ac{twem} is expected to associate different vectors to the word \textit{gay} at different times: its vector in $1900$ is expected to be more similar to the vector of \textit{joyful} than its vector in $2005$. By building a sequence of temporal embeddings of a word over consecutive time intervals, one can track the semantic shift occurred in the word usage. Moreover, temporal word embeddings make it possible to find distinct words that share a similar meaning in different periods of time, e.g., by retrieving temporal embeddings that occupy similar regions in the vector spaces that correspond to distinct time periods. 

The training process of a \ac{twem} relies on \textit{diachronic text corpora}, which are obtained by partitioning text corpora into temporal ``slices''~\cite{Hamilton2016DiachronicChange,Yao2017DiscoveryLearning}. Because of the stochastic nature of the neural networks training process, if we apply a Word2vec-like model on each slice, the output vectors of each slice will be placed in a vector space that has a different coordinate system.
This will preclude comparison of vectors across different times \cite{Hamilton2016DiachronicChange}. A close analogy would be to ask two cartographers to draw a map of Italy during different periods, without giving either of them a compass: the maps would be similar, although one will be rotated by an unknown angle with respect to the other~\cite{Smith2017ODictionary}. To be able to compare embeddings across time, 
their vector spaces corresponding to different time periods have to be aligned.  

Most of the proposed \ac{twem}s align multiple vector spaces by enforcing word embeddings in different time periods to be similar \cite{Kulkarni2015StatisticallyChange,Rudolph2017DynamicEvolution}. 
This method is based on the assumption that the majority of the words do not change their meaning over time.  
This approach is well motivated but may lead, for some words, to excessively smoothen differences between meanings that have shifted along time. 
A remarkable limitation of current \ac{twem}s is related to the assumptions they make on the size of the corpus needed for training: while some methods like~\cite{Szymanski2017TemporalEmbeddings,Hamilton2016DiachronicChange} require a huge amount of training data, which may be difficult to acquire in several application domains, other methods like~\cite{Yao2017DiscoveryLearning,Rudolph2017DynamicEvolution} may not scale well when trained with big datasets.

In this work we propose a new heuristic to train temporal word embeddings that has two main objectives: 1) to be simple enough to be executed in a \textit{scalable and efficient} manner, thus easing the adoption of temporal word embeddings by a large community of scientists, 
and 2) to produce models that achieve \textit{good performance} when trained with both small and big datasets. 

The proposed heuristic exploits the often overlooked dual representation of words that is learned in the two Word2vec architectures: Skip-gram and Continuous bag-of-words (CBOW) \cite{Mikolov2013EfficientSpace}. Given a target word, Skip-gram tries to predict its contexts, i.e., the words that occur nearby the target word (where ``nearby'' is defined using a window of fixed size); given a context, CBOW tries to predict the target word appearing in that context. 
In both architectures, each word is represented by a \textit{target embedding} and a \textit{context embedding}.  
During training, the target embedding of each word is placed nearby the context embeddings of the words that usually appear inside its context.
Both kinds of vectors can be used to represent the word meanings~\cite{Jurafsky2006SpeechRecognition}.

The heuristic consists in keeping one kind of embeddings frozen across time, e.g., the target embeddings, and using a specific temporal slice to update the other kind of embeddings, e.g., the context embeddings.
The embedding of a word updated with a slice corresponds to its temporal word embedding relative to the time associated with this slice.
The frozen embeddings act as an \textit{atemporal compass} and makes sure that the temporal embeddings are already generated during the training inside a shared coordinate system. In reference to the analogy of the map drawing ~\cite{Smith2017ODictionary}, our method draws maps according to a compass, i.e., the reference coordinate system defined by the atemporal embeddings.

In a thorough experimental evaluation conducted using temporal analogies and held-out tests, we show that our approach outperforms or equals comparable state-of-the-art models in all the experimental settings, despite its efficiency, simplicity and increased robustness against the size of the training data.  
The simplicity of the training method and the interpretability of the model (inherited from Word2vec) may foster the application of temporal word embeddings in studies conducted in related research fields, similarly to what happened with Word2vec which was used, e.g., to study biases in language models \cite{caliskan2017semantics}.    

The paper is organized as follows: in Section 2 we summarize the most recent approaches on temporal word embeddings. In Section 3 we present our model while in Section 4 we present experiments with state-of-the-art datasets. Section 5 ends the paper with conclusions and future work.

\section{Related Work}\label{relatedwork}
Different researchers have investigated the use of word embeddings to analyze semantic changes of words over time~\cite{Hamilton2016DiachronicChange,Kulkarni2015StatisticallyChange}. 
We identify two main categories of approaches based on the strategy applied to align temporal word embeddings associated with different time periods. 

\emph{Pairwise alignment}-based approaches align pairs of vector spaces to a unique coordinate system: \citeauthor{Kim2014TemporalModels} and \citeauthor{DelTredici2016TracingSpaces} align consecutive temporal vectors through neural network initialization; other authors apply various linear transformations after training that minimize the distance between the pairs of vectors associated with each word in two vector spaces \cite{Kulkarni2015StatisticallyChange,Hamilton2016DiachronicChange,Szymanski2017TemporalEmbeddings,Zhang2016TheTime}. 

\emph{Joint alignment}-based approaches train all the temporal vectors concurrently, enforcing them inside a unique coordinate system:  \citeauthor{Bamman2014DistributedLanguage} extend Skip-gram Word2vec tying all the temporal embeddings of a word to a common global vector (they originally apply this method to detect geographical language variations); other models impose constraints on consecutive vectors in the PPMI matrix factorization process \cite{Yao2017DiscoveryLearning} or when training probabilistic models to enforce the ``smoothness'' of the vectors' trajectory along time~\cite{Bamler2017DynamicEmbeddings,Rudolph2017DynamicEvolution}. This strategy leads to better embeddings when smaller corpora are used for training but is less efficient then pairwise alignment. 

Despite the differences, both alignment strategies try to enforce the vector similarity among different temporal embeddings associated with a same word. 
While this alignment principle is well motivated from a theoretical and practical point of view, enforcing the vector similarity of one word across time may lead to excessively smooth the differences between its representations in different time periods. Finding a good trade-off between \textit{dynamism} and \textit{staticness} seems an important feature of a \ac{twem}.      
Finally, very few models proposed in the literature
do not require explicit pairwise or joint alignment of the vectors, and they all rely on co-occurrence matrix or high-dimensional vectors \cite{Gulordava2011ACorpus,Basile2016DiachronicNgram}. 

In our work we present a neural model that use the same assumption of the ones proposed in the literature but does not require explicit alignment between different temporal word vectors. The main contribution of the model proposed in this paper with respect to state-of-the-art \ac{twem}s is to jointly offer the following main features: 
(i) to implicitly align different temporal representations using a shared coordinate system instead of enforcing (pairwise or joint) vector similarity in the alignment process; (ii) to rely on neural networks and low-dimensional word embeddings; (iii) to be easy to implement on top of the well-known Word2vec and highly efficient to train.

\section{\acl{atmodel}}

\begin{figure}
\centering
\includegraphics[width=0.95\columnwidth]{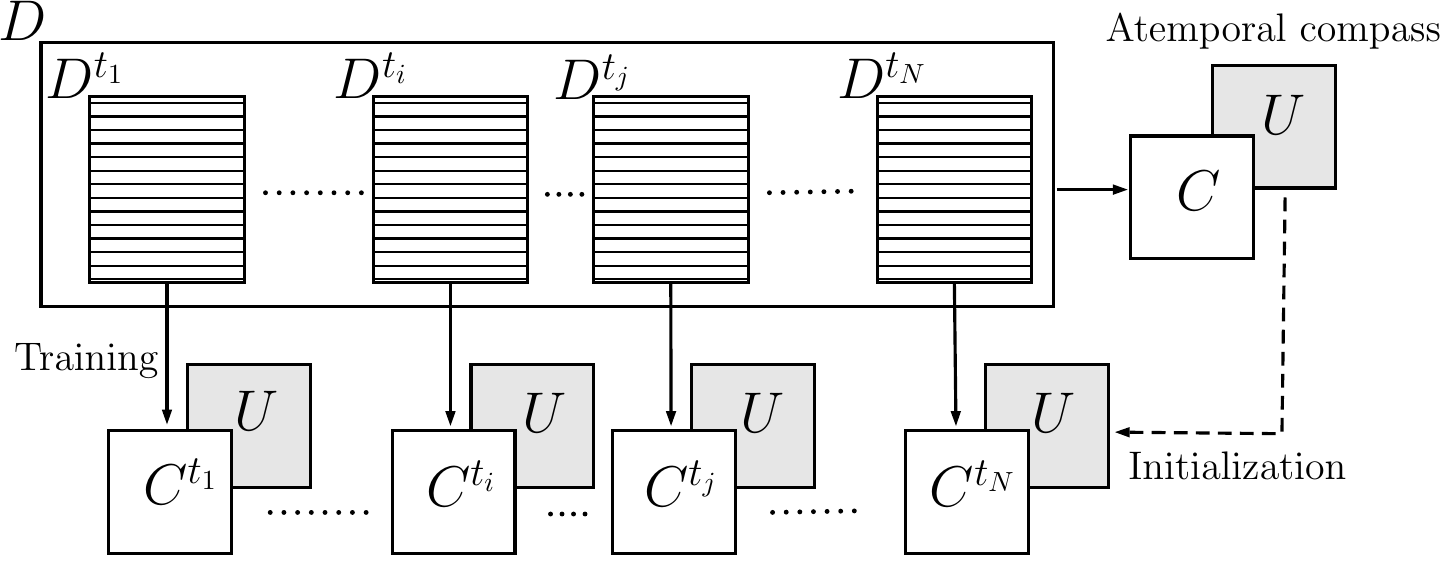}
\caption{The \ac{atmodel} model. The temporal context embeddings $\mathbf{C}^t$ are independently trained over each temporal slice, with frozen pre-trained atemporal target embeddings $\mathbf{U}$.}\label{fig:twem}
\end{figure}

We refer the \ac{twem} introduced in this paper as \textit{Temporal Word Embeddings with a Compass} (\ac{atmodel}), because of the compass metaphor used for their training. 

Our approach is based on the same assumption used in previous work, that is, the majority of words do not change their meaning over time \cite{Kulkarni2015StatisticallyChange}.
From this assumption, we derive a second one: we assume that a shifted word, i.e., a word whose meaning has shifted across time,  appears in the contexts of words whose meaning changes  slightly. However, differently from the latter assumption, our assumption is particularly true for shifted words. For example, the word \textit{clinton} appears during some temporal periods in the contexts of words that are related to his position as president of the USA (e.g., \textit{president}, \textit{administration}); conversely, the meanings of these words have not changed. 
The above assumption allows us to heuristically consider the target embeddings as static, i.e., to freeze them during training, while allowing the context embeddings to change based on co-occurrence frequencies that are specific to a given temporal interval. 
Thus, our training method returns the context embeddings as temporal word embeddings. 
 
Finally, we observe that our compass method can be applied also in the opposite way, i.e., by freezing the context embeddings and moving the target embeddings, which are eventually returned as temporal embeddings. However, a thorough comparison between these two specular compass-based training strategies is out of the scope of this paper.

\ac{atmodel} can be implemented on top of the two Word2vec models,  Skip-gram and CBOW. 
 
Here we present the details of our model using CBOW as underlying Word2vec model, since we empirically found that it produces temporal models that show better performance than Skip-gram with small datasets. We leave to the reader the interpretation of our model when Skip-gram is used. 

In the CBOW model, context embeddings $\vec{c}_j$ are encoded in the \textit{input weight matrix} $\mathbf{C}$ of the neural network, while target embeddings $\vec{u}_k$ are encoded inside the \textit{output weight matrix} $\mathbf{U}$ (vice-versa in the Skip-gram model).
Let us consider a diachronic corpus $D$ divided in $n$ temporal slices $D^{t_i}$, with $1\leq i\leq n$. 
The training process of \ac{atmodel} is divided in two phases, which are schematically depicted in Figure \ref{fig:twem}. 

(i) First, we construct two \textit{atemporal} matrices $\mathbf{C}$ and $\mathbf{U}$ by applying the original CBOW model on the whole diachronic corpus $D$, ignoring the time slicing; $\mathbf{C}$ and $\mathbf{U}$ represents the set of \textit{atemporal context embeddings} and \textit{atemporal target embeddings}, respectively. 
(ii) Second, for each time slice $D^{t_i}$, we construct a temporal context embedding matrix $\mathbf{C}^{t_i}$ as follows. We initialize the output weight matrix of the neural network with the previously trained target embeddings from the matrix $\mathbf{U}$. We run the CBOW algorithm using the temporal slice $D^{t_i}$. During this training process, the target embeddings of the output matrix $\mathbf{U}$ are not modified, while we update the context embeddings in the input matrix $\mathbf{C}^{t_i}$. After applying this process on all the time slices $D^{t_i}$, each input matrix $\mathbf{C}^{t_i}$ will represent our temporal word embeddings at the time $t_i$. Here below we further explain the key phase in our model, that is, the update of the input matrix for each slice, and the interpretation of the update function in our temporal model. 

Given a temporal slice $D^{t}$, the second phase of the training process can be formalized for a single training sample $\langle w_{k},\gamma(w_{k}) \rangle \in D^{t}$ as the following optimization problem:
\begin{equation}
\begin{split}
\max_{\mathbf{C}^t} \log P(w_{k}|\gamma(w_{k})) =  \sigma(\vec{u}_{k} \cdot \vec{c}_{\gamma(w_{k})}^{\;t}) 
\end{split}\label{loss}
\end{equation}
where $\gamma(w_k)=\langle w_{j_1},\cdots,w_{j_M} \rangle$ represents the $M$ words in the context of $w_k$ which appear in $D^t$ ($\frac{M}{2}$ is the size of the context window), $\vec{u}_k \in \mathbf{U}$ is the atemporal target embedding of the word $w_k$, and 
\begin{equation}
\begin{split}
\vec{c}_{\gamma(w_{k})}^{\;t} = \frac{1}{M} (\vec{c}_{j_1}^{\;t} + \cdots + \vec{c}_{j_M}^{\;t})^T\\
\end{split}\label{c_cw}
\end{equation}
is the mean of the temporal context embeddings $\vec{c}_{j_m}^{\;t}$ of the contextual words $w_{j_m}$. The softmax function $\sigma$ is calculated using Negative Sampling \cite{Mikolov2013DistributedCompositionality}. Please note that $\mathbf{C}^t$ is the only weight matrix to be optimized in this phase ($\mathbf{U}$ is constant), which is the main difference from the classic CBOW. The training process maximizes the probability that given the context of a word $w_k$ in a particular temporal slice $t$, we can predict that word using the atemporal target matrix $\mathbf{U}$. Intuitively, it moves the temporal context embedding $\vec{c}_{j_m}^{\;t}$ closer to the atemporal target embeddings $\vec{u}_{k}$ of the words that usually have the word $w_{j_m}$ in their contexts during the time $t$. The resulting temporal context embeddings can be used as temporal word embeddings: they will be already aligned, thanks to the shared atemporal target embeddings used as a compass during the independent trainings.

The proposed method can be viewed as a method that implements the main intuition of \cite{Gulordava2011ACorpus} using neural networks, and as a simplification of the models of \cite{Rudolph2017DynamicEvolution,Bamman2014DistributedLanguage}. Despite this simplification, experiments show that \ac{atmodel} outperforms or equals more sophisticated version on different settings. 
Our model has the same complexity of CBOW over the entire corpus $D$, plus the task of computing $n$ CBOW models over all the time slices. 

We observe that, differently from those approaches that enforce similarity between consecutive word embeddings (e.g., \citeauthor{Rudolph2017DynamicEvolution}), \ac{atmodel} does not apply any time-specific assumption. In the next section, we show that this feature does not affect the quality of the temporal word embeddings generated using \ac{atmodel}. Otherwise, this feature makes \ac{atmodel}'s training process more general and applicable to corpora sliced using different criteria, e.g., a news corpus split by location, publisher, or topic, to study differences in meaning that depend on factors other than time. 

\section{Experiments}\label{experiments}
In this section, we discuss the experimental evaluation of \ac{atmodel}. We compare \ac{atmodel} with static models and with the state-of-the-art temporal models that have shown better performance according to the literature. We use the two main methodologies proposed to evaluate temporal embeddings so far: temporal analogical reasoning~\cite{Yao2017DiscoveryLearning} and held-out tests~\cite{Rudolph2017DynamicEvolution}. Our experiments can be easily replicated using the source code available online\footnote{https://github.com/valedica/twec}. 

\subsection{Experiments on Temporal Analogies}

\noindent \textbf{Datasets} We use two datasets with different sizes to test the effects of the corpus size on the models' performances. The \textit{small dataset} \cite{Yao2017DiscoveryLearning} is freely available online\footnote{https://sites.google.com/site/zijunyaorutgers/publications}. 
We will refer to this dataset as \ac{datas}. The \textit{big dataset} is the New York Times Annotated Corpus\footnote{https://catalog.ldc.upenn.edu/ldc2008t19}  \cite{Sandhaus2008TheCorpus} employed by \citeauthor{Szymanski2017TemporalEmbeddings,Zhang2016TheTime} to test their TWEMs. We will refer to this dataset as \ac{datab}. Both datasets are divided into yearly time slices. 

We chose two test sets from those already available: \ac{tests} introduced by \citeauthor{Yao2017DiscoveryLearning} and \ac{testb} introduced by \citeauthor{Szymanski2017TemporalEmbeddings}.
They are both composed of temporal word analogies based on publicly recorded knowledge, partitioned in categories (e.g., \textit{President of the USA, Super Bowl Champions}). 
The main characteristics of datasets and test sets are summarized in Table \ref{test:tabledata}.

\begin{table}[]
\centering
\small
\begin{tabular}{|l|l|l|l|l|l|}
\hline  
Data  & Words &  Span & Slices   \\ \hline
\ac{datas}  &
$50$M &  $1990$-$2016$ & $27$ \\ \hline
\ac{datab}  & $668$M
 &  $1987$-$2007$ & $21$  \\ \hline
 \acs{dataml}  & $6.5$M &  $2007$-$2015$ & $9$
 \\ \hline \hline
Test &  Analogies &  Span & Categories \\ \hline
\ac{tests}  &
$11,028$ & $1990$-$2016$ & $25$ \\ \hline
\ac{testb}  &
 $4,200$ &  $1987$-$2007$ & $10$  \\ \hline
\end{tabular}%
\caption{Details of \ac{datas}, \ac{datab}, \acs{dataml}, \ac{tests} and \ac{testb}.}
\label{test:tabledata}
\end{table}

\noindent \textbf{Methodology}
To test the models trained on \ac{datas} we used the \ac{tests}, while to test the models trained on \ac{datab} we used the \ac{testb}. This allows us to replicate the settings of the work of \citeauthor{Yao2017DiscoveryLearning} and \citeauthor{Szymanski2017TemporalEmbeddings} respectively.
We quantitative evaluate the performance of a TWEM in the task of solving \textit{temporal word analogies} (TWAs) \cite{Szymanski2017TemporalEmbeddings}. The task of solving a temporal word analogy, given in the form $w_1 : t_1 = x : t_2$, is to find the word $w_2$ that is the most semantically similar word in $t_2$ to the \textit{input word} $w_1$ in $t_1$, where $t_1$ and $t_2$ are temporal intervals. Because semantically similar words result in distributional similar ones, it follows that two words involved in a TWA will occupy a similar position in the vector space at different points in time. Then, solving the TWA $w_1 : t_1 = x : t_2$ consists in finding the nearest vector at time $t_2$ to the input vector of the word $w_1$ at time $t_1$. For example, we expect that the vector of the word \textit{clinton} in $1997$ will be similar to the vector of the word \textit{reagan} in $1987$. 

Given an analogy $w_1 : t_1 = x : t_2$, we define \textit{time depth} $\delta_t$ as the distance between the temporal intervals involved in the analogy: $\delta_t=|t_1-t_2|$. 
Analogies can be divided in two subsets: the set $Static$ of \textit{static analogies}, which involve a pair of the same words (\textit{obama : 2009 = obama : 2010}), and the set $Dynamic$ $ (Dyn.)$ of \textit{dynamic analogies}, that are not static. We refer to the complete set of analogies as $All$. Given a \ac{twem} an a set of TWAs, the evaluation on the given answers is done with the use of two standard metrics, the Mean Reciprocal Rank (MRR) and 

\noindent \textbf{Algorithms}
We tested different models to compare the results of \ac{atmodel} with the ones provided by the state-of-the-art: two models that apply pairwise alignment, two models that apply joint alignment and a baseline static model. Where not stated differently, we implemented them with CBOW and Negative Sampling extending the \textit{gensim} library. The compare \ac{atmodel} with the following models:
\begin{itemize}
\item \textit{LinearTrans-Word2vec} (\textit{TW2V}) \cite{Szymanski2017TemporalEmbeddings}. 
\item \textit{OrthoTrans-Word2vec} (\textit{OW2V}) \cite{Hamilton2016DiachronicChange}.
\item \textit{Dynamic-Word2vec} (\textit{DW2V}) \cite{Yao2017DiscoveryLearning}. The original model was not available for replication at the time of the experiments. However, they provide the dataset and the test set of their evaluation settings (the same employed in our experiments) and published their results using our same metrics. Thus, the model can be compared to our using the same parameters.
\item \textit{Geo-Word2vec} (\textit{GW2V})
\cite{Bamman2014DistributedLanguage}. We use the implementation provided by the authors.
\item \textit{Static-Word2vec} (\textit{SW2V}): a baseline adopted by \citeauthor{Yao2017DiscoveryLearning} and \citeauthor{Szymanski2017TemporalEmbeddings}. The embeddings are learned over all the diachronic corpus, ignoring the temporal slicing.
\end{itemize}
We also tested the model defined by \citeauthor{Rudolph2017DynamicEvolution} and we obtained results close to the baseline SW2V, confirming what reported by \citeauthor{barranco2018tracking}; Thus, we do not report the results for their model on the analogy task.

\subsubsection{Experiments on \ac{datas}}
The first setting involves all the presented models, trained on \ac{datas} and tested over \ac{tests}. The hyper-parameters reflect those of \citeauthor{Yao2017DiscoveryLearning}: small embeddings of size $50$, a window of $5$ words, $5$ negative samples and a small vocabulary of $21$k words with at least $200$ occurrences over the entire corpus.  Table \ref{test:set1:table_accuracysd} summarizes the results.

\begin{figure}[]
\centering
\includegraphics[width=0.95\columnwidth]{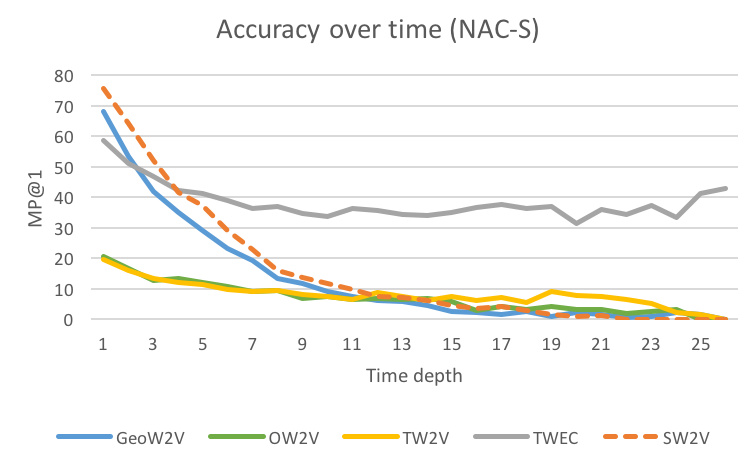}
\caption{Accuracy (MP@$1$) as function of time depth $\delta_t$ in \ac{tests}. Given an analogy $w_1 : w_2 = t_1 : t_2$, the time depth is plotted as $\delta_t=|t_1-t_2|$.}\label{test:set1:fig_accuracy}
\end{figure}

We can see that \ac{atmodel} outperforms the other models with respect to all the employed metrics. In particular, it performs better than DW2V, giving $7$\% more correct answers. 
DW2V confirms its superiority with respect to the pairwise alignment methods, as in \citeauthor{Yao2017DiscoveryLearning}. Unfortunately, due to the lack of the answers set and the embeddings, we can not know how well it performs over static and dynamic analogies separately. TW2V and OW2V scored below the static baseline (as in \citeauthor{Yao2017DiscoveryLearning}), particularly on analogies with small time depth (Figure \ref{test:set1:fig_accuracy}). In this setting, the pairwise alignment approach leads to huge disadvantages due to data sparsity: the partitioning of the corpus produces tiny slices (around $3.5$k news articles) that are not sufficient to properly train a neural network; the poor quality of the embeddings affects the subsequent pairwise alignment. As expected, SW2V's accuracy on analogies drops sharply as time depth increases (Figure \ref{test:set1:fig_accuracy}). On the contrary, \ac{atmodel}, TW2V and OW2V maintain almost steady performances over different time depths. GW2V does not answer correctly almost any dynamic analogies. We conclude that GW2V alignment is not capable of capturing the semantic dynamism of words across time for the analogy task. For this reason, we do not employ it in our second setting.

\begin{table}[]
\small
\centering
\begin{tabular}{|l|l|l|l|l|l|l|}
\hline  
Model & Set & MRR & MP1 & MP3 & MP5 & MP10\\ \hline  
\multirow{2}{*}{SW2V} &Static&$\textbf{1}$&$\textbf{1}$&$\textbf{1}$&$\textbf{1}$&$\textbf{1}$\\ \cline{2-7} 
                        &Dyn.&$0.148$&$0.000$&$0.263$&$0.351$&$0.437$\\ \cline{2-7}
                      &All&$0.375$&$0.266$&$0.459$&$0.524$&$0.587$\\ \hline
\multirow{2}{*}{TW2V} &Static&$0.245$&$0.193$&$0.280$&$0.313$&$0.366$\\ \cline{2-7} 
                        &Dyn.&$0.106$&$0.069$&$0.123$&$0.156$&$0.205$\\ \cline{2-7}
                      &All&$0.143$&$0.102$&$0.165$&$0.198$&$0.248$\\ \hline
\multirow{2}{*}{OW2V} &Static&$0.265$&$0.202$&$0.299$&$0.348$&$0.415$\\ \cline{2-7} 
                        &Dyn.&$0.087$&$0.058$&$0.099$&$0.124$&$0.160$\\ \cline{2-7}
                      &All&$0.135$&$0.096$&$0.153$&$0.183$&$0.228$\\ \hline
\multirow{2}{*}{DW2V} &Sta&$-$&$-$&$-$&$-$&$-$\\ \cline{2-7} 
                        &Dyn.&$-$&$-$&$-$&$-$&$-$\\ \cline{2-7} 
                      &All&$0.422$&$0.331$&$0.485$&$0.549$&$0.619$\\ \hline
\multirow{2}{*}{GW2V} &Static&$0.857$&$0.819$&$0.888$&$0.909$&$0.931$\\ \cline{2-7} 
                        &Dyn.&$0.071$&$0.005$&$0.092$&$0.159$&$0.225$\\ \cline{2-7}
                      &All&$0.280$&$0.222$&$0.305$&$0.359$&$0.435$\\ \hline \hline
\multirow{2}{*}{\ac{atmodel}} &Static&$0.720$&$0.668$&$0.763$&$0.787$&$0.813$\\ \cline{2-7} 
                        &Dyn.&$\textbf{0.394}$&$\textbf{0.308}$&$\textbf{0.451}$&$\textbf{0.508}$&$\textbf{0.571}$\\ \cline{2-7} 
                      &All&$\textbf{0.481}$&$\textbf{0.404}$&$\textbf{0.534}$&$\textbf{0.582}$&$\textbf{0.636}$\\ \hline
\end{tabular}

\caption{MRR and MP for the subsets of static and dynamic analogies of \ac{tests}. We use MPK in place of MP@K.
DW2V results are taken from the original paper \cite{Yao2017DiscoveryLearning}.}
\label{test:set1:table_accuracysd}
\end{table}

\begin{figure}
\centering
\includegraphics[width=.95\columnwidth]{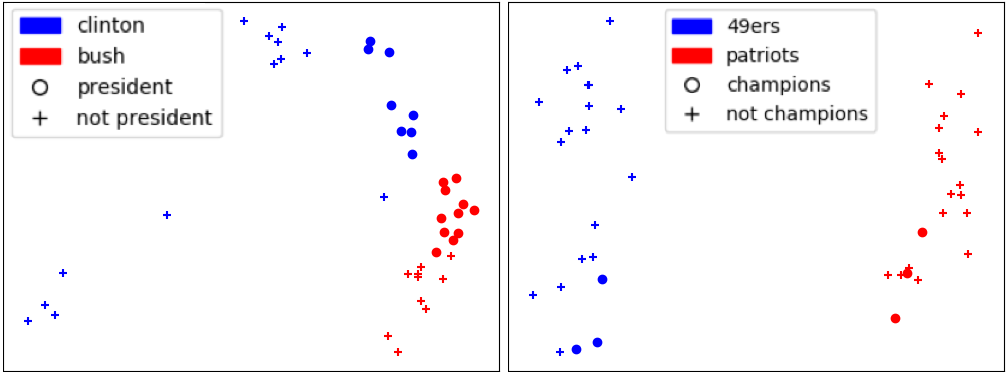}
\caption{2-dimensional PCA projection of the temporal embeddings of pairs of words from \textit{clinton}, \textit{bush} and \textit{49ers},  \textit{patriots}. The dot points highlight the temporal embeddings during their presidency or their winning years.}\label{qual:pairs_president}
\end{figure}

The comparison of the models' performances across the $25$ categories of analogies contained in \ac{tests} reveals new information: TW2V and OW2V's correct answers cover mainly $4$ categories, like \textit{President of the USA} and \textit{President of the Russian Federation}; \ac{atmodel} scores better over all the categories. Some categories are more difficult than others: even \ac{atmodel} scores nearly $0$\% in many categories, like \textit{Oscar Best Actor and Actress} and \textit{Prime Minister of India}. This discrepancy may be due to various reasons. First of all, some categories of words are more frequent than others in the corpus, so their embeddings are better trained. For example, \textit{obama} occurs $20,088$ times in \ac{datas}, whereas \textit{dicaprio} only $260$. As noted by \citeauthor{Yao2017DiscoveryLearning}, in the case of some categories of words, like presidents and mayors, we are heavily assisted by the fact that they commonly appear in the context of a title (e.g. \textit{President Obama}, \textit{Mayor de Blasio}). For example in \ac{atmodel}, \textit{obama} during its presidency is always the nearest context embedding to the word \textit{president}. Lastly, as noted by \citeauthor{Szymanski2017TemporalEmbeddings}, some roles involved in the analogies only influence a small part of an entity's overall news coverage. We show that this is reflected in the vector space: as we can see in Figure \ref{qual:pairs_president}, presidents' embeddings almost cross each other during their presidency, because they share a lot of contexts; on the other hand, football teams' embeddings remain distant. 

\subsubsection{Experiments on \ac{datab}}

This setting involves four models: SW2V, TW2V, OW2V, and \ac{atmodel}. The models are trained on \ac{datab} and tested over \ac{testb}. The settings' parameters are similar to those of \citeauthor{Szymanski2017TemporalEmbeddings}: longer embeddings of size $100$, a window size of $5$, $5$ negative samples and a very large vocabulary of almost $200$k words with at least $5$ occurrences over the entire corpus. Table \ref{test:set2:table_accuracysd} summarizes the results.

\begin{table}[]
\centering
\small
\begin{tabular}{|l|l|l|l|l|l|l|}
\hline  
Model & Set & MRR & MP$1$ & MP$3$ & MP$5$ & MP$10$\\ \hline 
\multirow{2}{*}{SW2V} &Static&$\textbf{1}$&$\textbf{1}$&$\textbf{1}$&$\textbf{1}$&$\textbf{1}$\\ \cline{2-7} 
                      &Dyn.&$0.102$&$0.000$&$0.149$&$0.259$&$0.326$\\ \cline{2-7} 
                        &All&$0.283$&$0.201$&$0.321$&$0.408$&$0.462$\\  \hline
\multirow{2}{*}{TW2V} &Static&$0.842$&$0.805$&$0.869$&$0.890$&$0.915$\\ \cline{2-7} 
                        &Dyn.&$0.343$&$0.287$&$0.377$&$0.414$&$0.467$\\ \cline{2-7}
                      &All&$0.444$&$0.391$&$0.476$&$0.510$&$0.558$\\ \hline
\multirow{2}{*}{OW2V} &Static&$0.857$&$0.824$&$0.876$&$0.903$&$0.926$\\ \cline{2-7} 
                        &Dyn.&$0.346$&$\textbf{0.290}$&$0.379$&$0.420$&$0.462$\\ \cline{2-7} 
                      &All&$0.449$&$0.398$&$0.480$&$0.518$&$0.556$\\ \hline \hline
\multirow{2}{*}{\ac{atmodel}} &Static&$0.948$&$0.936$&$0.959$&$0.961$&$0.967$\\ \cline{2-7} 
                        &Dyn.&$\textbf{0.367}$&$0.287$&$\textbf{0.423}$&$\textbf{0.471}$&$\textbf{0.526}$\\ \cline{2-7}  
                      &All&$\textbf{0.484}$&$\textbf{0.418}$&$\textbf{0.531}$&$\textbf{0.570}$&$\textbf{0.615}$\\ \hline
\end{tabular}
\caption{MRR and MP for the subsets of static and dynamic analogies of \ac{testb}. We use MPK in place of MP@K.}
\label{test:set2:table_accuracysd}
\end{table}

\ac{atmodel} still outperforms all the other models with respect to all the metrics, although its advantage is less than in the previous setting. Generally, we can say that \ac{atmodel} assigns lower ranks to correct answer words comparing to TW2V and OW2V; however it assigns the first position to them as frequently as the competing models. Table \ref{test:set2:table_accuracysd} shows that the advantage of \ac{atmodel} is limited to the static analogies. TW2V and OW2V score much better results than in the previous setting. This is due to the increased size of the input dataset which allows the training process to work well on individual slices of the corpus. 

\begin{figure}
\centering
\includegraphics[width=0.95\columnwidth]{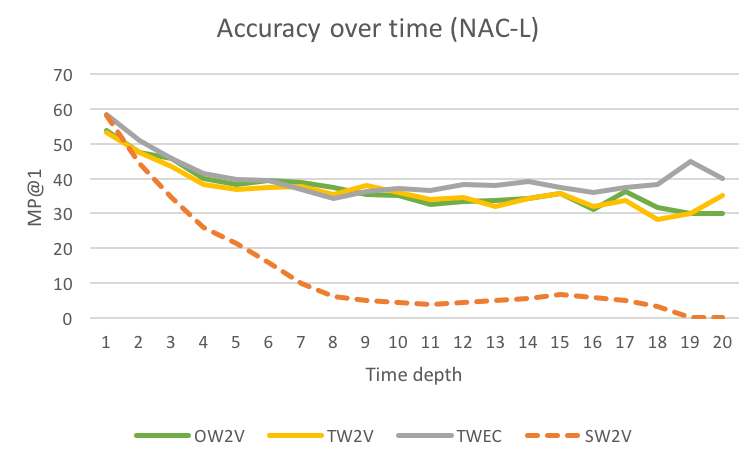}
\caption{Accuracy (MP@$1$) as function of time depth $\delta_t$ in \ac{testb}. Given an analogy $w_1 : w_2 = t_1 : t_2$, the time depth is plotted as $\delta_t=|t_1-t_2|$.}\label{test:set2:fig_accuracy}
\end{figure}

In Figure \ref{test:set2:fig_accuracy} we can see how the three temporal models behave similarly with respect to the time depth of the analogies.

The comparison of the models' performances across the $10$ categories of analogies contained in \ac{testb} reveals more differences between them. The results in terms of MP@$1$ are summarized in Table \ref{test:set2:accuracy_cat}. TW2V and OW2V significantly outperform \ac{atmodel} in two categories: \textit{President of the USA} and \textit{Super Bowl Champions}. In both cases, this is due to the major accuracy on dynamic analogies; most of the time, \ac{atmodel} is wrong because it gives static answers to dynamic analogies.
\ac{atmodel} significantly outperforms the other models in two categories: \textit{WTA Top-ranked Player} and \textit{Prime Minister of UK}. However in this case, \ac{atmodel} outperforms them both on dynamic and static analogies. 

\begin{table}[]
\centering
\small
\resizebox{.95\columnwidth}{!}{
\begin{tabular}{|l|l|l|l|l|}
\hline  
Category & SW2V & TW2V & OW2V & \ac{atmodel} \\ \hline
President of the USA&0.4000&\textbf{0.9905}&\textbf{0.9905}&0.8833\\ \hline
Secret. of State (USA)&0.1190&0.3000&\textbf{0.3619}&0.3405\\ \hline
Mayor of NYC&0.2476&\textbf{0.9643}&0.9524&0.9405\\ \hline
Gover. of New York&0.4476&0.9333&0.9381&\textbf{0.9786}\\ \hline
Super Bowl Champ.&0.0571&0.2024&\textbf{0.2524}&0.1452\\ \hline
NFL MVP&\textbf{0.0190}&0.0143&0.0143&\textbf{0.0190}\\ \hline
Oscar Best Actress&0.0095&\textbf{0.0119}&0.0071&\textbf{0.0119}\\ \hline
WTA Top Player&0.1619&0.2071&0.1548&\textbf{0.2857}\\ \hline
Goldman Sachs CEO&\textbf{0.1762}&0.0143&0.0238&0.1190\\ \hline
UK Prime Minister&0.3762&0.2762&0.2857&\textbf{0.4595}\\ \hline
\end{tabular}
}
\caption{Accuracy (MP@$1$) for the subsets of the analogy categories in \ac{testb}. The best scores are highlighted.}
\label{test:set2:accuracy_cat}
\end{table}

\subsection{Experiments on Held-Out Data}
In this section we show the performance of the \ac{atmodel} on an held-out task. We perform this test in two different ways. We tried to replicate the likelihood based experiments in~\citeauthor{Rudolph2017DynamicEvolution} and to further give confirmation about the performance of our model we also test the posterior probabilities using the framework described in~\citeauthor{taddy2015document}. 
Given a model, \citeauthor{Rudolph2017DynamicEvolution} assign a Bernoulli probability to the observed words in each held-out position: this metric is straightforward because it corresponds to the probability that appears in Equation \ref{loss}. However, at the implementation level, this metric is highly affected by the magnitude of the vectors because is based on the dot product of the vectors $\vec{u}_k$ and $\vec{c}_{\gamma(w_k)}$. In particular, \citeauthor{Rudolph2017DynamicEvolution} applied L2 regularization on the embeddings, which prioritize vectors with small magnitude.

This makes the comparison between models trained with different methods more difficult. Furthermore, we claim that held-out likelihood is not enough to evaluate the quality of a \ac{twem}: a good temporal model should be able to extract discriminative features from each temporal slice and to improve its likelihood based on them. To quantify this specific quality, we propose to adapt the task of document classification for the evaluation of \ac{twem}. We take advantage of the simple theoretical background and the easy implementation of the work of ~\citeauthor{taddy2015document}. We show that this new metric is not affected by the different magnitude of the compared vectors. 

\begin{figure*}[]
\centering
\includegraphics[width=0.95\textwidth]{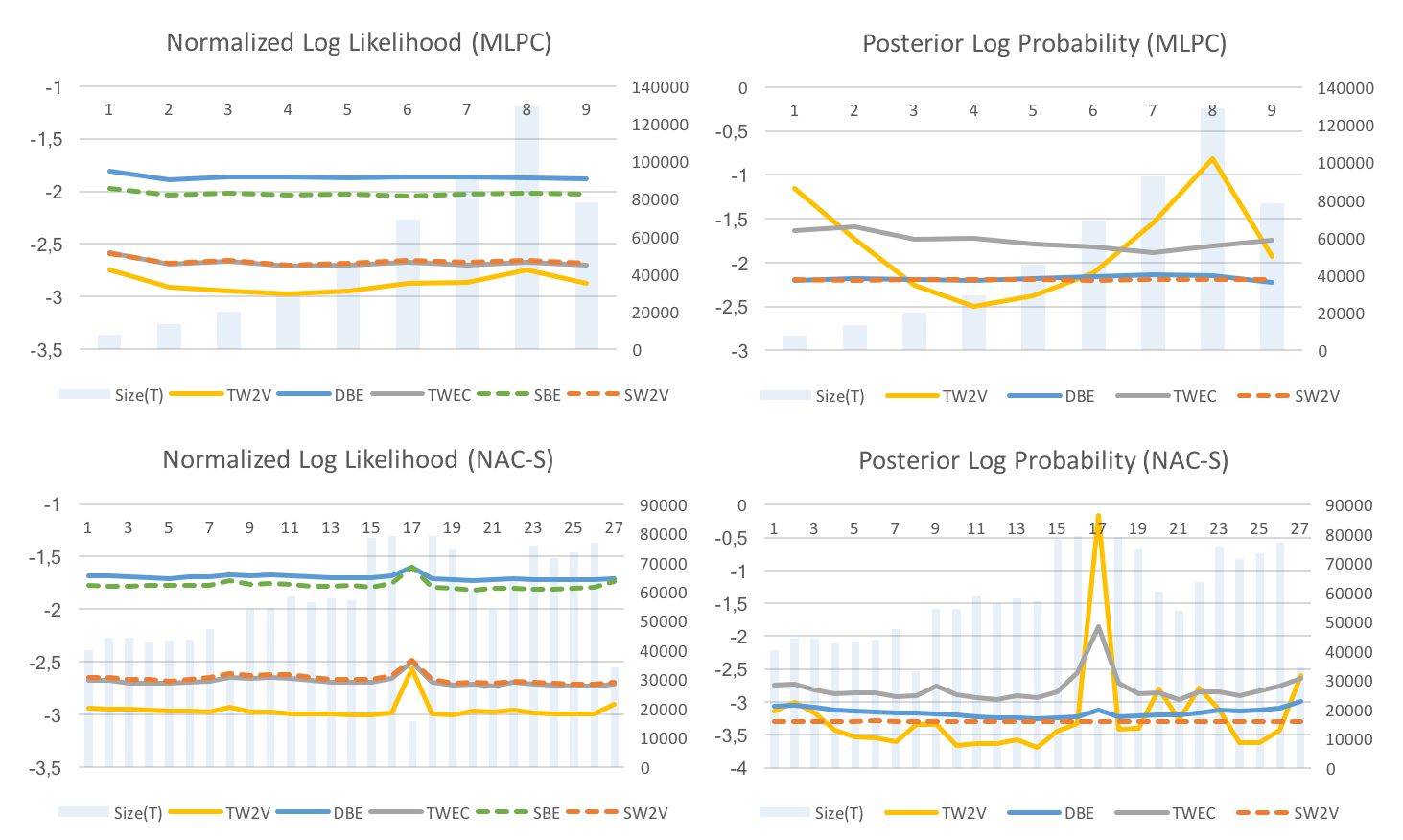}
\caption{$\mathcal{L}_{\mathcal{V}}^t$ and $\mathcal{P}_{\mathcal{V}}^t$ for each test slice $\mathcal{T}^t$ and model $\mathcal{V}$. Blue bars represent the number of words in each slice.}
\label{img:likelihood:and:posteriors} 
\end{figure*}

\noindent \textbf{Dataset}
We study two datasets, whose details are summarized in Table  \ref{test:tabledata}. The \ac{dataml} contains the full text from all the machine learning papers published on the ArXiv between $9$ years, from April 2007 to June 2015. The size of each slice is very small (less than $130,000$ words after pre-processing) and it increases over years. \ac{dataml} is made available online~\cite{Rudolph2017DynamicEvolution} by \citeauthor{Rudolph2017DynamicEvolution}: the text is already pre-processed, sub-sampled ($|V|=5,000$) and split into training, validation and testing ($80\%$, $10\%$, $10\%$); the dataset is shared in a computer-readable format without sentence boundaries: we convert it to plain text and we arbitrarily split it into 20-word sentences, suited to our training process. We also employ the \ac{datas} dataset, described in previous sections. Compared to the first dataset, \ac{datas} has $\times3$ more slices and it has approximately $60,000$ words per slice, with the exception of the year 2006. We use the same pre-processing script of \citeauthor{Rudolph2017DynamicEvolution} to prepare the \ac{datas} dataset for training and testing ($|V|=21,000$).

\noindent \textbf{Methodology A}
We measure the held-out likelihood following a methodology similar to \citeauthor{Rudolph2017DynamicEvolution}.   Given a \ac{twem} $ \mathcal{V}=\{\mathcal{V}_{t \in 1 \cdots T}\} = \{ \langle \mathbf{C}^{t \in 1 \cdots T},\mathbf{U}^{t \in 1 \cdots T}\rangle \} $, we calculate the log-likelihood for the temporal testing slice $\mathcal{T}^t=\langle w_1,\cdots,w_N \rangle$ as:
\begin{equation} \label{heldout_loglike}
\begin{split}
\log P_{\mathcal{V}_{t}}(\mathcal{T}^t) & = \sum_{n=1}^{N} \log P_{\mathcal{V}_{t}}(w_n|\gamma(w_n))
\end{split}
\end{equation}
where the probability $\log P_{\mathcal{V}_{t}}(w_n|\gamma(w_n))$ is calculated based on Equation \ref{loss} using Negative Sampling and the vectors of $\mathbf{C}^t$ and $\mathbf{U}^t$. As \citeauthor{Rudolph2017DynamicEvolution}, we equally balance the contribution of the positive and negative samples. For each model $\mathcal{V}$, we report the value of the normalized log likelihood $\mathcal{L}^t$:
\begin{equation} \label{norm_log_like}
\mathcal{L}_{\mathcal{V}}^t = \frac{1}{N} \log P_{\mathcal{V}}(\mathcal{T}^t)
\end{equation}
and its arithmetic mean $\mathcal{L}_{\mathcal{V}}$ over all the slices.

\noindent \textbf{Methodology B}
We adapt the methodology of \citeauthor{taddy2015document} to the evaluation of \ac{twem}. We calculate the posterior probability of assessing a temporal testing slice $\mathcal{T}^t$ to the correct temporal class label $t$. In our setting, this corresponds to the probability that a model $\mathcal{V}$ predicts the year of the $t$-th slice given an held-out text from the same slice. We apply Bayes rules to calculate this probability: 
\begin{equation} P_{\mathcal{V}_t}(t|\mathcal{T}^t)= \frac{P_{\mathcal{V}_t}(\mathcal{T}^t)P(t)}{\sum_{k=1}^{T}P_{\mathcal{V}_k}(\mathcal{T}^t)P(k)}
\end{equation}
A good temporal model $\mathcal{V}=\{\mathcal{V}_{t \in 1 \cdots T}\} $ will assign an high likelihood to the slice $\mathcal{T}^t$ using the vectors of $\mathcal{V}_t$ and a relatively low likelihood using the vectors of $\mathcal{V}_{k\neq t}$.
We assume that the prior probability on class label $t$ is the same for each class, $P(t)=1/T$. For implementation reason, we redefine the posterior likelihood as:
\begin{equation}
\mathcal{P}_\mathcal{V}^t=P_{\mathcal{V}_t}(t|\mathcal{T}^t)= \frac{1}{S} \sum_{s=1}^{S} \frac{P_{\mathcal{V}_t}({z}_s^t)P(t)}{\sum_{k=1}^{T} P_{\mathcal{V}_k}({z}_s^t)P(k)}
\end{equation}
where ${z}_s$ is the $s$-th sentence in $\mathcal{T}^t$ and $P_{\mathcal{V}_{t}}({z}_s)$ is calculated based on Equation \ref{heldout_loglike}. Please note that this metric is not affected by the magnitude of the vectors because is based on a ratio of probabilities. For each model $\mathcal{V}$, we report the value of the posterior log probability $\mathcal{P}_\mathcal{V}^t$ and its arithmetic mean $\mathcal{P}_\mathcal{V}$ over all the slices.

\noindent \textbf{Algorithms}
We test five temporal models for this setting: our model \ac{atmodel}, TW2V, the baseline SW2V, the \textit{Dynamic Bernoulli Embeddings} (DBE) \cite{Rudolph2017DynamicEvolution} and the \textit{Static Bernoulli Embeddings} (SBE) \cite{rudolph2016exponential}. Note that TW2V is equivalent to OW2V in this setting because we do not need to align vectors from different slices. DBE is the temporal extension of SBE, a probabilistic framework based on CBOW: it enforces similarity between consecutive word embeddings using a prior in the loss function, and specularly to \ac{atmodel}, it uses a unique representation of context embeddings for each word. We trained all the models on the temporal training slices $D^t$ using a CBOW architecture, a shared vocabulary and the same parameters, which are similar to \citeauthor{Rudolph2017DynamicEvolution}: learning rate $\eta=0.0025$, window of size $1$, embeddings of size $50$ and $10$ iterations ($5$ static and $5$ dynamic for \ac{atmodel}, $1$ static and $9$ dynamic for DBE as suggested by \citeauthor{Rudolph2017DynamicEvolution}). 
Following \citeauthor{Rudolph2017DynamicEvolution}, before the second phase of the training process of \ac{atmodel}, we initialize the temporal models with both the weight matrices $\mathbf{C}$ and $\mathbf{U}$ of the static model: we note that this improves held-out performances but it negatively affects the analogy tests. 
We limit our study to small datasets and small embeddings due to the computational cost: DBE takes almost $6$ hours to train on \ac{datas} on a $16$-core CPU setting. DBE and SBE are implemented by the authors using \textit{tensorflow}, while all the other models are implemented in \textit{gensim}: to evaluate them, we convert them to \textit{gensim} models, extracting the matrices $\mathbf{U}^t$ and $\mathbf{C}$.

\subsection{Results}
Table \ref{test:heldout} shows the mean results of the two metrics for each model. In both settings, \ac{atmodel} obtain a likelihood almost equal to SW2V but a much better posterior probability than the baseline. This is remarkable considering that \ac{atmodel} optimizes the scoring function only on one weight matrix $\mathbf{C}^t$, keeping the matrix $\mathbf{U}^t$ frozen. Comparing to TW2V, \ac{atmodel} has a better likelihood and its posterior probability is more stable across slices (Figure \ref{img:likelihood:and:posteriors}). 
The likelihood scores of DBE and SBE are highly influenced by the different magnitude of their vectors: we can quantify the contribution of the applied L2 regularization comparing the two static baseline SBE and SW2V. Differently from \ac{atmodel}, DBE slightly improves the likelihood with respect to its baseline. However, regarding the posterior probability, \ac{atmodel} outperforms DBE. Our experiments suggest an inverse correlation between the capability of generalization and the capability of extracting discriminative features from small diachronic datasets. Finally, experimental results show that \ac{atmodel} captures discriminative features from temporal slices without losing generalization power.

\begin{table}[]
\small
\centering
\begin{tabular}{|l|l|l|l|l|l|l|}
\hline  
Dataset & M & SW2V & SBE & TWEC & DBE & TW2V \\ \hline  
\multirow{2}{*}{\ac{dataml}} & $\mathcal{L}_{\mathcal{V}}$&-$2.67$&-$2.02$&-$2.68$&-$\textbf{1.86}$&-$2.88$\\ \cline{2-7} 
    
    &$\mathcal{P}_{\mathcal{V}}$&-$2.20$&-$2.20$&-$\textbf{1.75}$&-$2.18$&-$2.83$\\ \cline{2-7}
                       \hline
\multirow{2}{*}{\ac{datas}} &$\mathcal{L}_{\mathcal{V}}$&-$2.66$&-$1.77$&-$2.69$&-$\textbf{1.70}$&-$2.96$\\ \cline{2-7} 
                        &$\mathcal{P}_{\mathcal{V}}$&-$3.30$&-$3.30$&-$\textbf{2.80}$&-$3.16$&-$3.24$\\ \cline{2-7} \hline
    
\end{tabular}
\caption{The arithmetic mean of the log likelihood $\mathcal{L}_{\mathcal{V}}$ and of the posterior log probability $\mathcal{P}_{\mathcal{V}}$ for each model $\mathcal{V}$. Based on the standard error on the validation set, all the reported results are significant.}
\label{test:heldout}
\end{table}

\section{Conclusions and Future Work}
In this paper, we have presented a novel approach to train temporal word embeddings using atemporal embeddings as a compass. The approach is scalable and effective. While the idea of using an atemporal compass to align slices implicitly surfaced in previous work, we encode this principle into a training method based on neural networks, which makes it simpler and efficient.

Results of a comparative experimental evaluation based on 
datasets and methodologies used in previous work suggest that, despite its simplicity and efficiency, our approach builds models that are of equal or better quality than the ones generated with comparable state-of-the-art approaches. In particular, when compared to scalable models based on pairwise alignment strategies \cite{Hamilton2016DiachronicChange,Szymanski2017TemporalEmbeddings}, our model achieves better performance when trained on a limited size corpus and comparable performance when trained on a large corpus. 
At the same time, when compared to models based on joint alignment strategies~\cite{Yao2017DiscoveryLearning,Rudolph2017DynamicEvolution}, our model is more efficient or it obtains better performance even on a limited size corpus.

A possible future direction of our work is to test our temporal word embeddings as features in natural language processing tasks, e.g., named entity recognition, applied to historical text corpora. 
Also, we would like to apply the proposed model to compare word meanings along dimensions different from time, by slicing the corpus with appropriate criteria. For example, inspired by previous work by \citeauthor{caliskan2017semantics}, we plan to use word embeddings trained with articles published by different newspapers to compare word usage and investigate potential language biases.

Other partition criteria to test are by location and topic. Finally, an interesting future application may be in the field of automatic translation: as shown by~\cite{mikolov2013exploiting}, the alignment of two embedding spaces trained on bilingual corpora moves embeddings of similar words in similar positions inside the vector space. 

\section{Acknowledgements}
This research has been supported in part by EU H2020 projects EW-Shopp - Grant n. 732590, and EuBusinessGraph - Grant n. 732003.

\bibliography{biblio.bib}
\bibliographystyle{aaai}
\end{document}